\documentclass[runningheads]{llncs}
\usepackage{graphicx}
\usepackage{csquotes}
\usepackage{mwe}
\usepackage{subfig}
\usepackage{listings}
\usepackage{multirow}

\usepackage{color}

\definecolor{dkgreen}{rgb}{0,0.6,0}
\definecolor{gray}{rgb}{0.5,0.5,0.5}
\definecolor{mauve}{rgb}{0.58,0,0.82}

\begin{document}
\title{LioNets: Local Interpretation of Neural Networks through Penultimate Layer Decoding}
%
%
\author{Ioannis Mollas
\and
Nikolaos Bassiliades
\and
Grigorios Tsoumakas
}
\authorrunning{I. Mollas et al.}
%
\institute{Aristotle University of Thessaloniki, Thessaloniki  54124, Greece 
\email{\{iamollas,nbassili,greg\}@csd.auth.gr}}
\maketitle              
\begin{abstract}
Technological breakthroughs on smart homes, self-driving cars, health care and robotic assistants, in addition to reinforced law regulations, have critically influenced academic research on explainable machine learning. A sufficient number of researchers have implemented ways to explain indifferently any black box model for classification tasks. A drawback of building agnostic explanators is that the neighbourhood generation process is universal and consequently does not guarantee true adjacency between the generated neighbours and the instance. This paper explores a methodology on providing explanations for a neural network's decisions, in a local scope, through a process that actively takes into consideration the neural network's architecture on creating an instance's neighbourhood, that assures the adjacency among the generated neighbours and the instance.The outcome of performing experiments using this methodology reveals that there is a significant ability in capturing delicate feature importance changes.

\end{abstract}
\section{Introduction}
Explainable artificial intelligence is a fast-rising area of computer science. Most of the research in this area is currently focused on developing methodologies and libraries for interpreting machine learning models for two main reasons: a) increased use of black box machine learning models, such as deep neural networks, in safety-critical applications, such as self-driving cars, health care and robotic assistants, and b) radical law changes empowering ethics and human rights, which introduced the right of users to an explanation of machine learning models' decisions that concern them. 

Local Explanators are methods aiming to explain individual predictions of a particular model. LIME~\cite{Ribeiro2016quotWhyClassifier} is a state-of-the-art methodology that first constructs a local neighbourhood around a given new unlabeled instance, by perturbing the instance's features, and then trains a simpler transparent decision model to extract the features' importance. Subsequent model agnostic methods like Anchors~\cite{RibeiroAnchors:Explanations}, X-SPELLS~\cite{OrestisBSc} and LORE~\cite{Guidotti2018LocalSystems} focused on generating better neighbourhoods. 

This paper is concerned with generating better neighbourhoods too. However, it focuses on neural network models in particular, in contrast to the {\em model agnostic} local explanators mentioned in the previous paragraph that can work with any type of machine learning model. Our approach is inspired by the following observation: small changes at the input layer might lead to large changes at the penultimate layer of a (deep) neural network, based on which the final decision of the network is taken. We hypothesize that creating neighbourhoods at the penultimate layer of the neural network instead, could lead to better explanations. 

To investigate this intuitive research hypothesis, we introduce our approach, dubbed LioNets (Local Interpretation Of Neural nETworkS through penultimate layer decoding). LioNets constructs a local neighbourhood at the penultimate layer of the neural network and records the network's decisions for this neighbourhood. However, in order to build a transparent local explanator, we need to have input representations at the original input space. To achieve this, LioNets trains a decoder that learns to reconstruct the input examples from their representations at the penultimate layer of the neural network. Taking together, the neural network model and the decoder resemble an autoencoder.

For the evaluation of LioNets, a set of experiments have been conducted, whose code is available at GitHub repository ``LioNets''\footnote{\url{https://github.com/iamollas/LioNets}}. The results show that LioNets can lead to more precise explanations than LIME. 

\section{Background and Related Work}
In order to be able to present LioNets architecture, this section will provide a sequence of definitions concerning the matter of explainable machine learning, autoencoders and knowledge distillation.

\subsection{Explainable Machine Learning}
Explainable artificial intelligence is a broad and fast-rising field in computer science. Recent works focus on ways to interpret machine learning models. Thus, this paper will focus on explainable machine learning. An accurate definition is the following: 

\begin{displayquote}
``An interpretable system is a system where a user cannot only see but also study and understand how inputs are mathematically mapped to outputs. This term is favoured over ``explainable'' in the ML context where it refers to the capability of understanding the work logic in ML algorithms'' \cite{Adadi2018PeekingXAI}.
\end{displayquote}

There are several dimensions that can define an interpretable system according to~\cite{Guidotti2018AModels}. One interesting dimension is the {\em scope} of interpretability. There are two different scopes. An interpretable system can provide global or/and local explanations for its predictions. Global explanations can present the structure of the whole system, while local explanations are focused on particular instances.

In the same paper, they are also presenting the desired features of any interpretable system. Those are:

\begin{itemize}
    \item \textbf{Interpretability}: Interpretability measures how much comprehensible is an explanation. In fact, there is not a formal metric because for every problem we measure different attributes.
    \item \textbf{Accuracy}: The accuracy, and probably other metrics, of the original model and the accuracy of the explanator.
    \item \textbf{Fidelity}: Fidelity describes the mimic ability of the explanator, namely the ability of the explanator on providing the same results as the model it explains for specific instances.
\end{itemize}

\subsection{Autoencoders}
Autoencoders is a growing area within deep learning~\cite{Liu2017AApplications}. An autoencoder is an unsupervised learning architecture and can be expressed as a function 
\begin{equation}
    {\displaystyle \textstyle f:X\rightarrow X}. 
\end{equation}
Autoencoder networks are widely used for reducing the dimensionality of the input data. They initially encode the original data into some latent representation and subsequently reconstruct the original data by decoding this representation to the original dimensions. The most common varieties of autoencoders are the three following:
\begin{itemize}
    \item \textbf{Vanilla}: A three-layered neural network with one hidden layer.
    \item \textbf{Multilayer}: A deeper neural network with more than one hidden or recurrent layers. For example Variational Autoencoders~\cite{kingma2013auto,rezende2014stochastic}.
    \item \textbf{Convolutional}: Used for image or textual data. In practice, the hidden layers are not fully connected, but convolutional layers.
\end{itemize}

\subsection{Related Work}
As already mentioned, LIME~\cite{Ribeiro2016quotWhyClassifier} is a state-of-the-art method for explaining predictions. It follows a simple pipeline. It generates a neighbourhood of a specific size for an instance by choosing randomly to put a zero value in one or more features of every neighbour. Then the cosine similarity of each neighbour with the original instance is measured and multiplied by one hundred. This constitutes the weight on which the simple linear model will depend on for its training. Thus, the most similar neighbours will have more impact on the training process of the linear model. A disadvantage of LIME is in sparse data. Due to the perturbation method that takes place on the original space, LIME can only generate $2^n$ different neighbours, where $n$ the number of non-zero values. For example, in textual data, in a sentence of six words represented as a vector of four thousand features, where each feature corresponds to a word from the vocabulary, the non-zero features are only six. Hence, only $2^6=64$ different neighbours can be generated. However, LIME will create a neighbourhood of five thousand instances by randomly sampling through the 64 unique neighbours.

X-SPELLS~\cite{OrestisBSc} is a forthcoming solution providing model agnostic local explanations to black boxes dealing with sentiment analysis problems. The core idea of this work is to generate neighbourhoods for instances, which they will contain semantically correct synthetic neighbours, using techniques similar to paraphrasing. By creating such neighbourhoods, using variational autoencoders~\cite{kingma2013auto,rezende2014stochastic} to create new examples in the latent space, the goal is to present some of these neighbours to the user as the explanation. To accomplish this, they train a decision tree on the neighbourhood with labels assigned from the black box and subsequently they are extracting the exemplars. 

Another set of methodologies in explaining decision systems, and specifically neural networks, are using Knowledge Distillation~\cite{Hinton2015DistillingNetwork,Frosst2017DistillingTree}. Those methods are trying to explain globally the whole structure and the predictions of a deep neural network, by distilling its knowledge to a transparent system. This idea originates by the Dark Knowledge Distillation~\cite{Sadowski2015DeepMatter}, which is trying to enhance the performance of a shallow network (the student) through the knowledge of a deeper and more complex network (the teacher). 

\section{LioNets}
\begin{figure}
\includegraphics[width=1\textwidth]{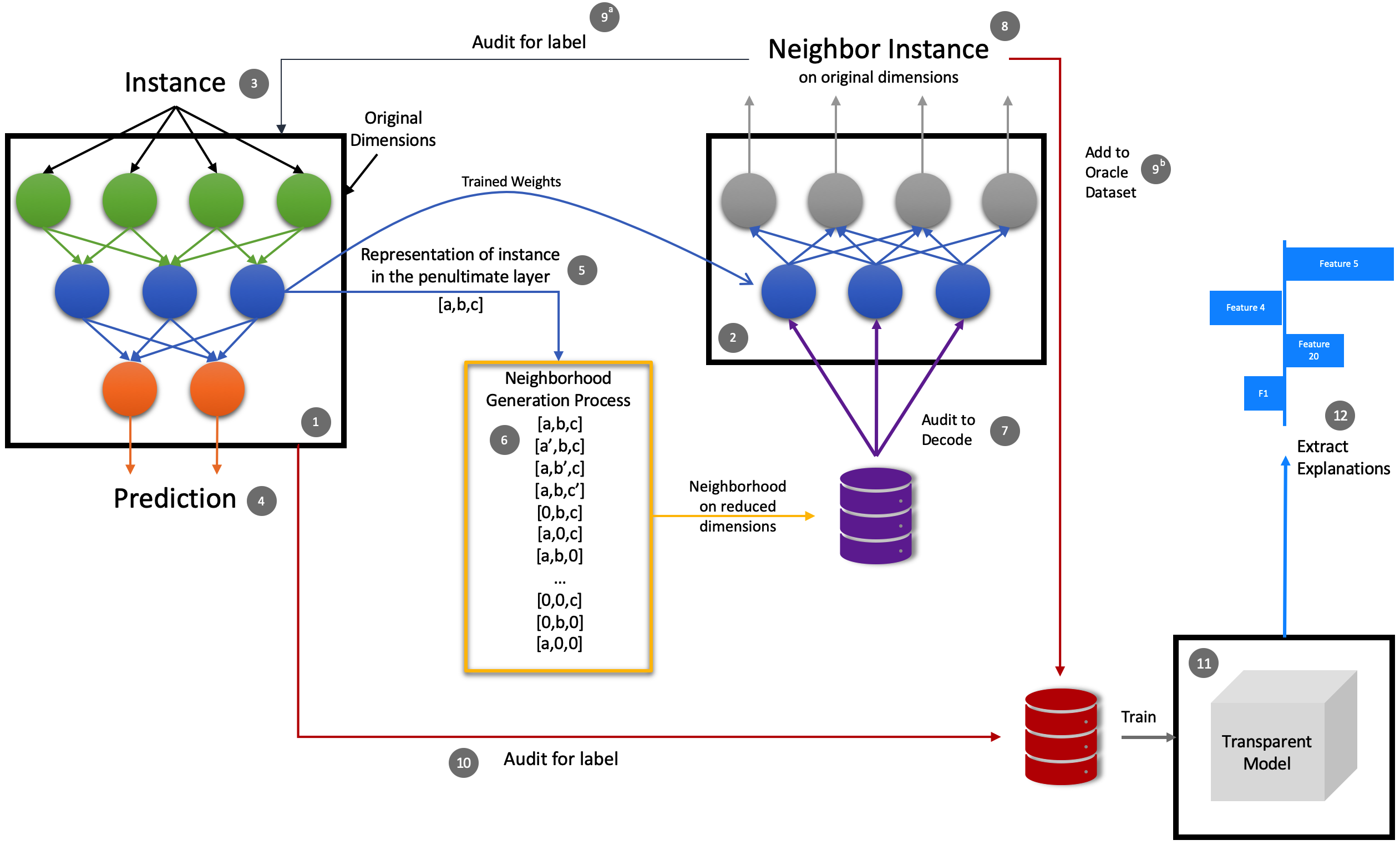}
\caption{LioNets' architecture. In this flow chart, the four fundamental mechanisms of LioNets are visible. In point 1 there is the predictor, while in point 2 the decoder. In point 3 there is the neighbourhood generation process and in point 4 the transparent model.} \label{fig1}
\end{figure}

This section presents the full methodology and architecture of LioNets. LioNets consist of four fundamental sub-architectures, which are visible in Fig~\ref{fig1} at points 1, 2, 6 and 11. The main part of such system is the neural network, which will work as the predictor. A decoder based on the predictor is the second part. Finally, a neighbourhood generation process and a transparent predictor are the last two mechanisms. Hence, the following process should be executed.

\subsection{Neural Network Predictor}
For a given dataset, a neural network with a suitable fine-tuned architecture is being trained on this dataset. The output layer is by design in the same length as the number of classes of the classification problem. This process is similar to other supervised methodologies of building and training a neural network for classification tasks, which defines a function ${\displaystyle \textstyle f:X\rightarrow Y}$.

\subsection{Encoder and Decoder}
When the training process of the neural network is over, a duplicate it is created. Then removing the last layer of this copy model and labelling every other layer as untrainable, the foundations for the autoencoder have been defined. Actually, these foundations would be the encoder, the first half of the autoencoder, thus only the decoder part is missing. By building successfully the decoder part and training it, the first two stages for Lionets' completion are achieved. Although, this is the most difficult stage to complete since it is not easy to successfully train autoencoders, especially when the first half of the autoencoder is untrainable. Another approach is to build the autoencoder firstly and afterwards to extract the layers in order to create the encoder, decoder and predictor networks.

Mathematically those neural networks can be expressed via these functions:
\begin{equation}
    Encoder {\displaystyle \textstyle:X\rightarrow Z},
\end{equation}
\begin{equation}
    Decoder {\displaystyle \textstyle:Z\rightarrow X},
\end{equation}
\begin{equation}
    Autoencoder {\displaystyle \textstyle :X\rightarrow X},
\end{equation}
\begin{equation}
    Predictor {\displaystyle \textstyle :X\rightarrow Y}.
\end{equation}

By keeping the encoder part untrainable with stable weights, it guarantees that the generated neighbourhood is transforming from the reduced dimensions to the original dimensions with a decoder, which was trained with the original architecture of the neural network. That process will produce a more representative neighbourhood for the instance, without any semantic meaning to humans.

The academic community has extensively explored ways to create better neighbourhoods for an instance, but every methodology was focused on generating new instances in the level of the input. In this work, the generation processes take place to the latent representation of the encoded input.

\subsection{Neighbourhood Generation Process}
The neighbourhood generation process takes place after the training of the neural network, the encoder and the decoder. This process could be a genetic algorithm, like the proposed methods in LORE~\cite{Guidotti2018LocalSystems} or even another neural network, but simpler solutions are preferred. In LioNets for an instance, that is desirable to get explanations, after encoding it via the encoder neural network, extracting its new representation form from the penultimate level of the neural network, the neighbourhood generation process begins with input the instance with reduced dimensions. By making small changes in the reduced space it could affect more than one dimensions of the original space. Thus, the simple feature perturbation methods on low dimensions will lead to a complex generated neighbour, which most probably would have no semantic meaning for humans.

At that point in time, a specific number of neighbours is generated through a selected generation process and that set of neighbours is given to the decoder, in order to be reversed to the original dimensions. 

\subsection{Transparent Predictor}
By the end of the neighbourhood generation stage, the neighbourhood dataset is almost complete. The only missing part is the neighbours' labels. Thus, the neural network is predicting each instance of the neighbourhood dataset assigning labels to every neighbour, in the form of probabilities. Afterwards, the final dataset with the neighbours and their labels are given as training data to any transparent regression model. The ultimate goal is to overfit that model to the training data.

\section{Evaluation}
The following section is presenting the setup for the experiments. The data preprocessing methods for two different datasets are described, alongside with the neural network models preparation and the neighbourhood generation process. Finally, there is a discussion about the results of the experiments.

\subsection{Setup}

Our experiments involve two textual binary classification datasets. The first one concerns the detection of hateful YouTube comments\footnote{https://intelligence.csd.auth.gr/research/hate-speech-detection}~\cite{Anagnostou2017Hatebusters:Speech} and contains 120 hate and 334 non-hate comments. The second dataset deals with the detection of spam SMS messages~\cite{Almeida2011ContributionsResults} and contains 747 spam and 4.827 ham (non-spam) messages. The pre-processing of these datasets consists of the following steps for each document:
\begin{itemize}
    \item Lowercasing,
    \item Stemming and Lemmatisation through WordNet lemmatizer \cite{miller1995wordnet} and Snowball stemmer \cite{snowball},
    \item Phrases transformations [Table~\ref{tab:my-tableTransform}],
	\item Removal of punctuation marks,
	\item Once again, Stemming and Lemmatisation.
\end{itemize}

\begin{table}
\centering
\begin{tabular}{ccc}
``what's''  & to & ``what is''  \\
``don't''   & to & ``do not''   \\
``doesn't'' & to & ``does not'' \\
``that's''  & to & ``that is''  \\
``aren't''  & to & ``are not''  \\
``'s''      & to & `` is''    \\
``isn't''   & to & ``is not''   \\
``\%''      & to & `` percent''  \\
``e-mail''  & to & ``email''    \\
``i'm''     & to & ``i am''     \\
``he's''    & to & ``he is''    \\
``she's''   & to & ``she is''   \\
``it's''    & to & ``it is''    \\
``'ve''     & to & `` have''    \\
``'re''     & to & `` are''     \\
``'d''      & to & `` would''   \\
``'ll''     & to & `` will''    
\end{tabular}
\caption{Phrases and words transformations.}
\label{tab:my-tableTransform}
\end{table}

Then, for transforming the textual data to vectors a simple term frequency-inverse document frequency \cite{sparck1972statistical} (TF-IDF) vectorization technique is taking place.

\begin{figure}

\begin{minipage}{.45\linewidth}
\centering
\subfloat[]{\label{main3:a}\includegraphics[scale=0.25]{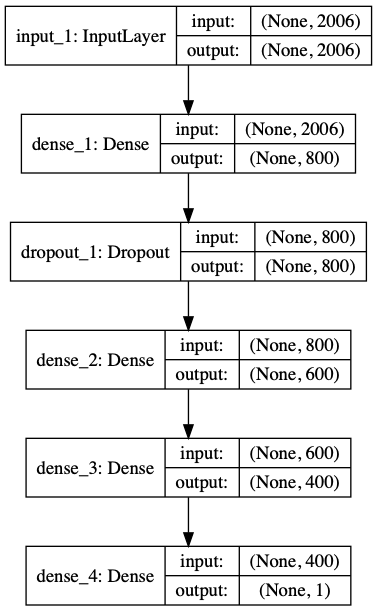}}
\end{minipage}%
\begin{minipage}{.45\linewidth}
\centering
\subfloat[]{\label{main3:b}\includegraphics[scale=0.37]{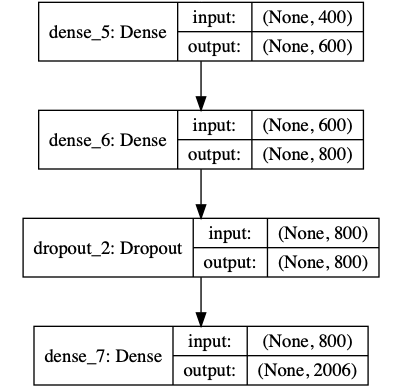}}
\end{minipage}\par\medskip
\caption{The predictor's architecture(a) and the decoder's architecture(b).}
\label{fig:main3}
\end{figure}

Afterwards, the neural network predictor for these experiments consists of six layers [Fig~\ref{main3:a}] and it has `binary\_crossentropy' as loss function. The encoder has five layers, which we extract from the predictor and the decoder has four layers as well [Fig~\ref{main3:b}], which we train using `categorical\_crossentropy' loss function. The autoencoder is the combination of the encoder and the decoder.

In this set of experiments, a simple generation process via features perturbation methods is applied. Specifically, the creation of neighbours for an instance emerges by multiplying one feature value at a time with $0$ and $2^z$, $z \in \{-2,-1,1,2\}$. Concisely, the above process generates instances which are different in only one dimension in their latent representation.

As soon as the neighbourhood is acquired, every neighbour is transformed via the decoder to the original dimensions. Then, the transformed neighbourhood is given as input to the predictor to predict the class probabilities. Finally, combining the output of the predictor with the transformed neighbourhood a new oracle dataset has been created. 
\\
\begin{lstlisting}[caption={Oracle dataset synthesis},captionpos=b]
    input: neighbourhood
    output: X, y
    transformed_neighbourhood = decoder.predict(neighbourhood)
    class_probabilities = predictor.predict(transformed_neighbourhood)
    X = transformed_neighbourhood, y = class_probabilities
\end{lstlisting}

The last step is to train a transparent model with this oracle dataset. It might be useful to check the distribution of probabilities of this dataset and if needed to transform it to have a normal distribution. In these experiments, the transparent model chosen is a Ridge Regression model. By training this model, the coefficients of the features are extracted and transformed into explanations, presented as features' weights in the x-axis of the following figures.

\begin{lstlisting}[caption={Explaining an instance},captionpos=b]
    input: X, y, instance, feature_names
    transparent_model = Ridge().fit(X,y)
    coef = transparent_model.coef_
    plot_explanation(coef*instance, feature_names)
\end{lstlisting}

\subsection{Results on the Hate Speech Dataset}

We take the following YouTube comment from the hate speech dataset as an example: ``aliens really, Mexicans are people too''. The true class of this comment is {\em no hate}. According to the neural network, the probability of the {\em hate} class is approximately 0.00208.

\begin{figure}[ht]
\begin{minipage}{.5\linewidth}
\centering
\subfloat[]{\label{main2:a}\includegraphics[scale=0.38]{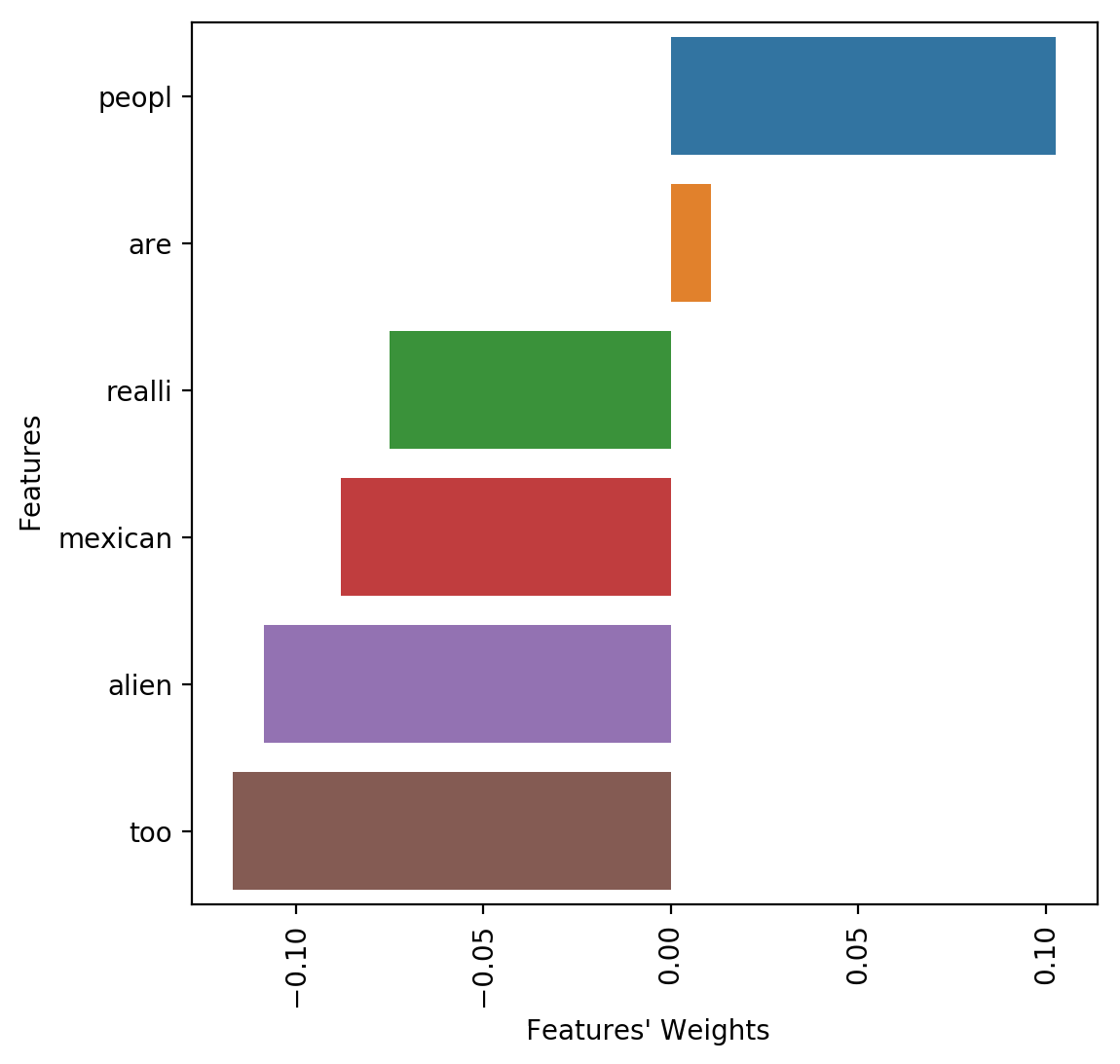}}
\end{minipage}%
\begin{minipage}{.5\linewidth}
\centering
\subfloat[]{\label{main2:b}\includegraphics[scale=0.38]{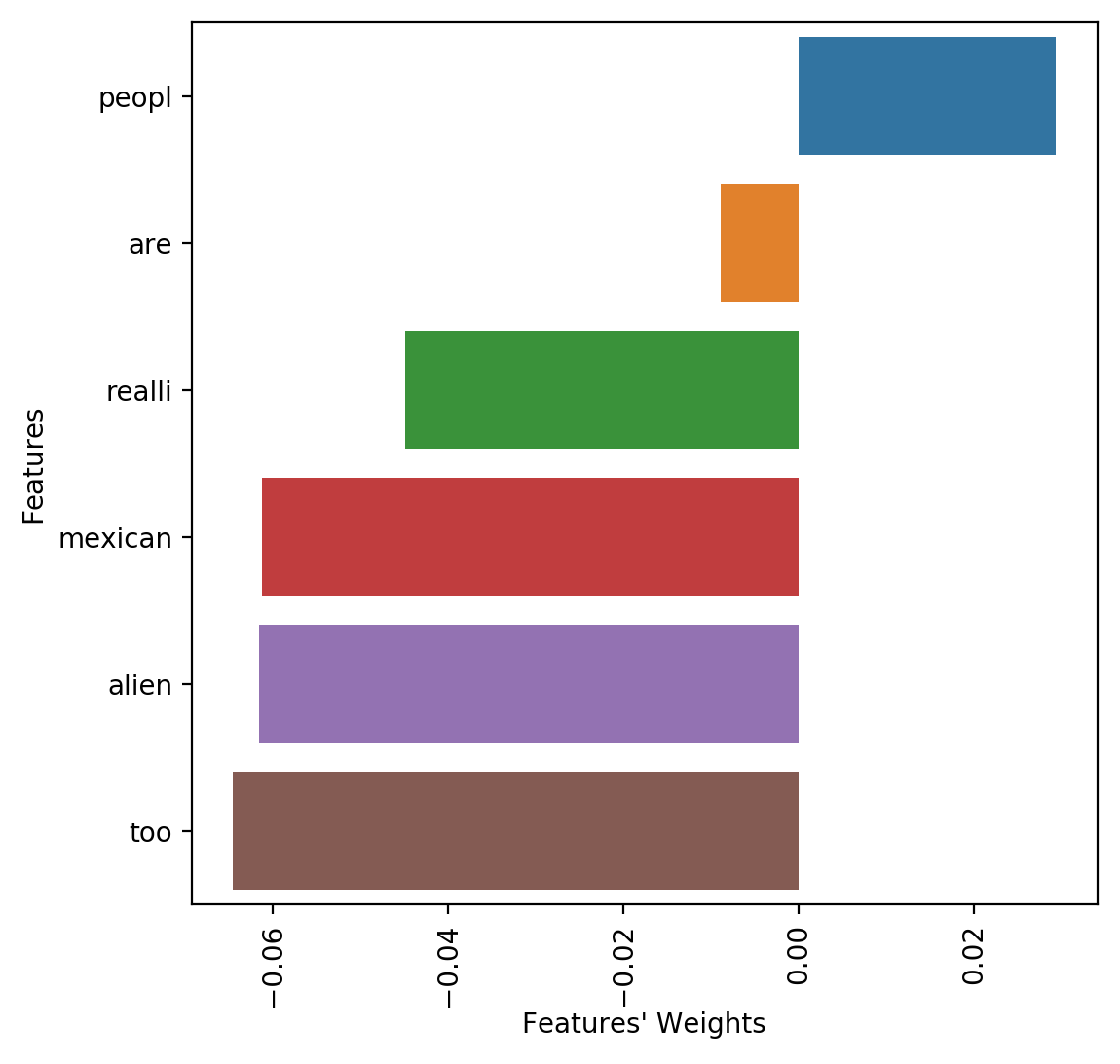}}
\end{minipage}\par\medskip
\caption{Explanation plots of a hate speech instance via (a) LioNets and (b) LIME.}
\label{fig:main2}
\end{figure}

Fig.~\ref{fig:main2} visualizes the explanation of the neural network's decision via LioNets (\ref{fig:main2}a) and LIME (\ref{fig:main2}b). At first sight, they appear similar. Their main difference is that they assign the feature's ``are'' contribution to different classes. By removing this word from the instance the neural network predicts 0.00197, which is a lower probability. Thus, it is clear that the feature ``are'' it was indeed contributing to the ``Hate Speech'' class for this specific instance as LioNets explained.

Although to support LioNets explanations, the generated neighbourhoods' distances from the original instance computed and presented in Table~\ref{tab:my-table1}. As it seems the neighbours generated by LIME on original space, in this example, when are encoded to the reduced space are further to the neighbours generated by LioNets in the encoded space. However, when the LioNets' generated neighbours are transformed back to the original space, are more distant to the original instance in comparison to LIME's neighbours, but that is the assumption that has been made through the beginning of these experiments. It is critical to mention at this point, that these distances measured with neighbours generated by changing only one feature at a time. 

\begin{table}[ht]
\centering
\begin{tabular}{|c|c|}
\hline
                                    & Euclidean distance \\ \hline
LIME: Generated on Original Space   & 0.3961             \\ \hline
LIME: Encoded                       & 0.9444             \\ \hline
LioNets: Generated on Encoded Space & 0.2163             \\ \hline
LioNets: Decoded to Original Space  & 0.7635             \\ \hline
\end{tabular}
\caption{Neighbourhood distances for instance of hate speech dataset.}
\label{tab:my-table1}
\end{table}

\subsection{Results on SMS spam dataset}

The second example which is going to be explained belongs to the SMS spam dataset. The text of the preprocessed instance is the following: ``Wife.how she knew the time of murder exactly''. This instance has true class ``ham''. The classifier predicted truthfully 0.00014 probability to be ``spam''.

\begin{figure}[ht]
\begin{minipage}{.5\linewidth}
\centering
\subfloat[]{\label{main:a}\includegraphics[scale=0.38]{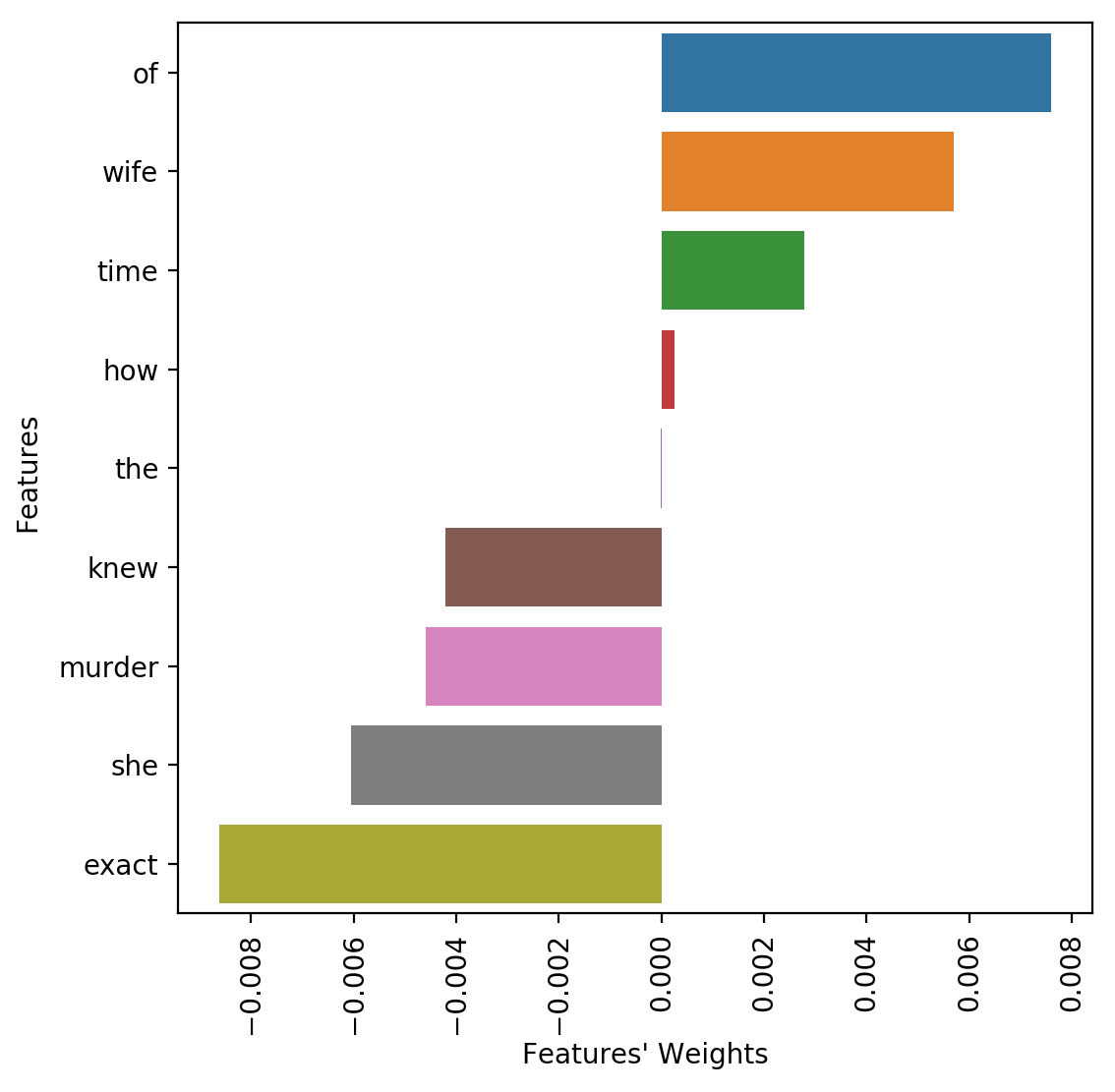}}
\end{minipage}%
\begin{minipage}{.5\linewidth}
\centering
\subfloat[]{\label{main:b}\includegraphics[scale=0.38]{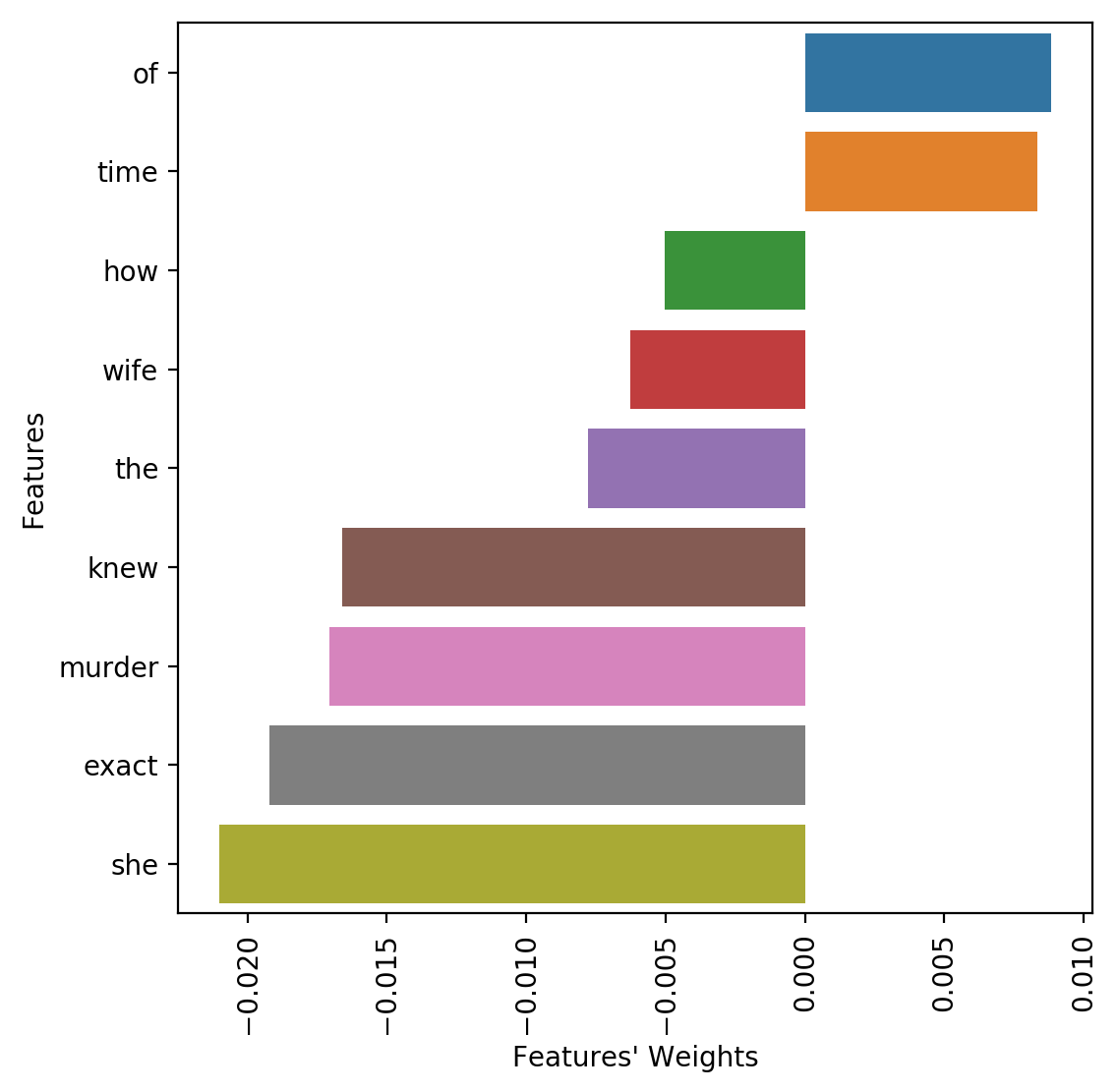}}
\end{minipage}\par\medskip
\caption{Explanation plots of SMS spam instance using LioNets(a) and LIME(b).}
\label{fig:main}
\end{figure}

Fig.~\ref{fig:main} presents two different explanations for the classifier's prediction. As before, Fig.~\ref{main:a} shows the explanation provided by LioNets and Fig.~\ref{main:b} shows LIME's explanation. The contribution of feature ``wife'' to the prediction is assigned to different classes in each explanation. To prove the stability and robustness of LioNets, this feature is removed and by auditing again the neural network the new prediction is lower with a probability of 0.000095. Thus, it is clear that feature ``wife'' was indeed contributing to the ``spam'' class as LioNets explained and captured.

Like before, the neighbourhoods' distances from the original instance are computed and presented in Table~\ref{tab:my-table2}. As it seems the neighbours generated by LIME on original space, by projecting them into the encoded space, are more distant to the encoded instance, compared to the neighbours generated by LioNets directly in the encoded space.

\begin{table}[ht]
\centering
\begin{tabular}{|c|c|}
\hline
                                    & Euclidean distance \\ \hline
LIME: Generated on Original Space   & 0.3184             \\ \hline
LIME: Encoded                       & 0.8068             \\ \hline
LioNets: Generated on Encoded Space & 0.3459             \\ \hline
LioNets: Decoded to Original Space  & 0.7875             \\ \hline
\end{tabular}
\caption{Neighbourhood distances for instance of SMS spam  collection.}
\label{tab:my-table2}
\end{table}

\section{Conclusion}
In summary, the LioNets architecture provides valid explanations for the decisions of a neural network that are comparable to other state-of-the-art techniques, while at the same time it guarantees better adjacency between the generated neighbours of an instance because the generation of the neighbours is performed on the penultimate layer of the network. In addition, LioNets can create better, larger and more representative neighbourhoods, because the generation process takes place at the encoded space, where the instance has a dense representation. These are the main points of creating and using LioNets on decision systems like neural networks. 

One main disadvantage of LioNets is that it is focused only on explaining neural networks, thus it is not a model agnostic method. Moreover, the overall process of building LioNets is harder than training neural network predictors, because they demand the training of a decoder, which is a difficult task.

Future work plans include testing the LioNets methodology on different variations of encoders and decoders and implementing more complex neighbourhood generation and neighbours selection processes. In addition, we would like to explore different transparent models for explaining the instances, such as rule-based models~\cite{Clark1989}, decision tree models~\cite{quinlan2014c4,breiman1984classification} and models based on abstract argumentation~\cite{DUNG1995321}. Lastly, we plan to evaluate LioNets based on human subject experiments.

\section*{Acknowledgment}
This paper is supported by the European Union's Horizon 2020 research and innovation programme under grant agreement No 825619. AI4EU Project\footnote{\url{https://www.ai4eu.eu}}.

\bibliographystyle{splncs04}
\bibliography{bib2}

\end{document}